\documentclass[10pt,letterpaper]{article}

\usepackage[top=0.85in,left=2.75in,footskip=0.75in,marginparwidth=2in]{geometry}
\usepackage[utf8]{inputenc}
\usepackage{cite}
\usepackage{nameref,hyperref}
\usepackage[right]{lineno}
\usepackage{microtype}
\DisableLigatures[f]{encoding = *, family = * }

\raggedright
\usepackage{changepage}
\usepackage[aboveskip=1pt,labelfont=bf,labelsep=period,singlelinecheck=off]{caption}

\makeatletter
\renewcommand{\@biblabel}[1]{\quad#1.}
\makeatother

\usepackage{lastpage,fancyhdr,graphicx}
\usepackage{epstopdf}
\fancyheadoffset[L]{2.25in}
\fancyfootoffset[L]{2.25in}

\usepackage{color}

\definecolor{Gray}{gray}{.25}

\usepackage{graphicx}

\usepackage{sidecap}

\usepackage{wrapfig}
\usepackage[pscoord]{eso-pic}
\usepackage[fulladjust]{marginnote}
\reversemarginpar

\begin{document}
\vspace*{0.35in}

\begin{flushleft}
{\Large
\textbf\newline{Doc2Vec on the PubMed corpus: study of a new approach to generate related articles}
}
\newline
\\
Emeric Dynomant\textsuperscript{1,2,4,*},
Stéfan J.~Darmoni\textsuperscript{1,3,4},
Émeline Lejeune\textsuperscript{1},
Gaëtan Kerdelhué\textsuperscript{1},
Jean-Philippe Leroy\textsuperscript{1},
Vincent Lequertier\textsuperscript{2},
Stéphane Canu\textsuperscript{4},
Julien Grosjean\textsuperscript{1,3}
\\
\bigskip
\bf{1} Department of Biomedical Informatics, University Hospital, Rouen, France.
\\
\bf{2} OmicX, Seine Innopolis, Le Petit Quevilly, Rouen, France.
\\
\bf{3} LIMICS, INSERM U1142, Sorbonne Université, Paris, France.
\\
\bf{4} LITIS, EA 4108, INSA, Rouen, France.
\\
\bigskip
* emeric.dynomant@omictools.com

\end{flushleft}

\section*{Abstract}
\textbf{Background} \hspace{1cm}
PubMed is the biggest and most used bibliographic database worldwide, hosting more than 26M biomedical publications. One of its useful features is the “similar articles” section, allowing the end-user to find scientific articles linked to the consulted document in term of context. The aim of this study is to analyze whether it is possible to replace the statistic model PubMed Related Articles (\textit{pmra}) with a document embedding method.\\
\textbf{Methods} \hspace{1cm}
Doc2Vec algorithm was used to train models allowing to vectorize documents. Six of its parameters were optimised by following a grid-search strategy to train more than 1,900 models. Parameters combination leading to the best accuracy was used to train models on abstracts from the PubMed database. Four evaluations tasks were defined to determine what does or does not influence the proximity between documents for both Doc2Vec and \textit{pmra}.\\
\textbf{Results} \hspace{1cm}
The two different Doc2Vec architectures have different abilities to link documents about a common context. The terminological indexing, words and stems contents of linked documents are highly similar between \textit{pmra} and Doc2Vec PV-DBOW architecture. These algorithms are also more likely to bring closer documents having a similar size. In contrary, the manual evaluation shows much better results for the \textit{pmra} algorithm.\\
\textbf{Conclusions} \hspace{1cm}
While the \textit{pmra} algorithm links documents by explicitly using terminological indexing in its formula, Doc2Vec does not need a
prior indexing. It can infer relations between documents sharing a similar indexing, without any knowledge about them, particularly regarding the PV-DBOW architecture. In contrary, the human evaluation, without any clear agreement between evaluators, implies future studies to better understand this difference between PV-DBOW and \textit{pmra} algorithm.



\section{Background}

\subsection{PubMed}

PubMed is the largest database of bio-medical articles worldwide with more than 29,000,000 freely available abstracts. Each article is identified by an unique PubMed IDentifier (PMID) and is indexed with the Medical Subject Headings (MeSH) terminology. In order to facilitate the Information Retrieval (IR) process for the end-user, PubMed launched in 2007 a service of related articles search, available both through its Graphical User Interface (GUI) and its Application Programming Interface (API). Regarding the GUI, while the user is reading a publication, a panel presents title of articles that may be linked to the current reading. For the API, the user must query eLink with a given PMID \cite{sayers2009utilities}. The output will be a list of others PMIDs, each associated with the similarity score computed by the \textit{pmra} (pubmed related article) model \cite{lin2007pubmed}.

\subsection{The \textit{pmra} model}

To do so, each document is tokenized into many topics $S_{i}$. Then, the probability $P(C|D)$ that the user will find relevant the document C when reading the document D will be calculated. For this purpose, the authors brought the concept of \textit{eliteness}. Briefly, a topic $S_{i}$ is presented as elite topic for a given document if a word $W_{i}$ representing $S_{i}$ is used with a high frequency in this document. This work allows to bring closer documents sharing a maximum of elite topics. In the article presenting the \textit{pmra} model, authors claim that “\textit{the deployed algorithm in PubMed also takes advantage of MeSH terms, which we do not discuss here}”. We can thus assume that a similar score is computed thanks to the associated MeSH terms with both documents D and C. Such an indexing is highly time-consuming and has to be manually performed.

\subsection{Documents embedding}

Nowadays, embedding models allow to represent a text into a vector of fixed dimensions. The primary purpose of this mathematical representation of documents was to be able to use texts as input of deep neural networks. However, these models have been used by the IR community as well: once all fitted
in the same multidimensional space, the cosine distance between two documents vectors can estimate the proximity between these two texts. In 2013, Mikolov \textit{et al.} released a word embedding method called Word2Vec (W2V) \cite{mikolov2013distributed}. Briefly, this algorithm uses unsupervised learning to train a model which embeds a word as a vector while preserving its semantic meaning. Following this work, Mikolov and Le released in 2014 a method to vectorize complete texts \cite{le2014distributed}. This algorithm, called Doc2Vec (D2V), is highly similar to W2V and comes with two architectures. The Distributed Memory Model of Paragraph Vectors (PV-DM) first trains a W2V model. This word embedding will be common for all texts from a given corpus C on which it was trained. Then, each document $D_{x}$ from C will be assigned to a randomly initialised vector of fixed length, which will be concatenated with vectors of words composing $D_{x}$ during the training time (words and documents vectors are sharing the same number of dimensions). This concatenation will be used by a final classifier to predict the next token of a randomly selected window of words. The accuracy of this task can be calculated and used to compute a loss function, used to back-propagate errors to the model, which leads to a modification of the document’s representation. The Distributed Bag of Words version of Paragraph Vector (PV-DBOW) is highly similar to the PV-DM, the main difference being the goal of the final classifier. Instead of concatenating vector from the document with word vectors, the goal here is to output words from this window just by using the mathematical representation of the document.

\subsection{Related Work}

Doc2Vec has been used for many cases of similar document retrieval. In 2016, Lee \textit{et al.} used D2V to clusterize positive and negative sentiments with an accuracy of 76.4\% \cite{lee2016sentiment}. The same year, Lau and Baldwin showed that D2V provides a robust representation of documents, estimated with two tasks: document similarity to retrieve 12 different classes and sentences similarity scoring \cite{lau2016empirical}. Recently, studies started to use documents embedding on the PubMed corpus. In 2017, Gargiulo \textit{et al.} used a combination of words vectors coming from the abstract to bring closer similar documents from Pubmed \cite{gargiulo2018deep}. Same year, Wang and Koopman used the PubMed database to compare D2V and their own document embedding method \cite{wang2017semantic}. Their designed accuracy measurement task was consisting in retrieving documents having a small cosine distance with the embedding of a query. Recently, Chen \textit{et al.} released BioSentVec, a set of sentence vectors created from PubMed with the algorithm sent2vec \cite{chen2018biosentvec,pagliardini2017unsupervised}. However, their evaluation task was based on public sentences similarity datasets, when the goal here is to embed entire abstracts as vectors and to use them to search for similar articles versus the \textit{pmra} model. In 2008, the related articles feature of PubMed has been compared (using a manual evaluation) with one that uses both a TF-IDF \cite{salton1988term} representation of the documents and Lin’s distance \cite{lin1998information} to compare their MeSH terms \cite{merabti2008searching}. Thus, no study was designed so far to compare documents embedding and the \textit{pmra} algorithm. The objectives of this study were to measure the ability of these two models to infer the similarity between documents from PubMed and to search what impacts the most this proximity. To do so, different evaluation tasks were defined to cover a wide range of aspects of document analogy, from their context to their morphological similarities.

\section{Methods}

\subsection{Material}

During this study, the optimisation of the model’s parameters and one of the evaluation tasks require associated MeSH terms with the abstracts from PubMed. Briefly, the MeSH is a medical terminology, used to index documents on PubMed to perform keywords-based queries. The MEDOC program was used to create a MySQL database filled with 26,345,267 articles from the PubMed bulk downloads on October 2018, 5\textit{th} \cite{dynomant2017medoc}. Then, 16,048,372 articles having both an abstract and at least one associated MeSH term were selected for this study. For each, the PMID, title, abstract and MeSH terms were extracted. The titles and abstracts were lowered, tokenized and concatenated to compose the PubMed documents corpus.

\subsection{Optimisation}

Among all available parameters to tune the D2V algorithm released by Gensim, six of them were selected for optimisation \cite{rehurek2010software}. The \textit{window\_size} parameter affects the size of the sliding window used to parse texts. The \textit{alpha} parameter represents the learning rate of the network. The \textit{sample} setting allows the model to reduce the importance given to high-frequency words. The \textit{dm} parameter defines the training used architecture (PV-DM or PV-DBOW). The \textit{hs} option defines whether hierarchical softmax or negative sampling is used during the training. Finally, the \textit{vector\_size} parameter affects the number of dimensions composing the resulting vector.

A list of possible values was defined for each of these six parameters. The full amount of possible combinations of these parameters were sent to slave nodes on a cluster, each node training a D2V model with a unique combination of parameters on 85\% of 100,000 documents randomly selected from the corpus. Every article from the remaining 15\% were then sent to each trained model and queried for the top-ten closest articles. For each model, a final accuracy score represented by the average of common MeSH terms percentage between each document $D_{i}$ from the 15,000 extracted texts and their returning top-ten closest documents was calculated. The combination of parameters with the highest score was kept for both PV-DBOW and PV-DM.

\subsection{Training}

The final models were trained on a server powered by four XEON E7 (144 threads) and 1To of RAM. Among the total corpus (16,048,372 documents), 1\% (160,482) was extracted as a test set (named \textbf{TeS}) and was discarded from the training. The final models were trained on 15,887,890 documents representing the training set called \textbf{TrS}.

\subsection{Evaluation}

\begin{figure}
 \centering
 \includegraphics[scale=0.3]{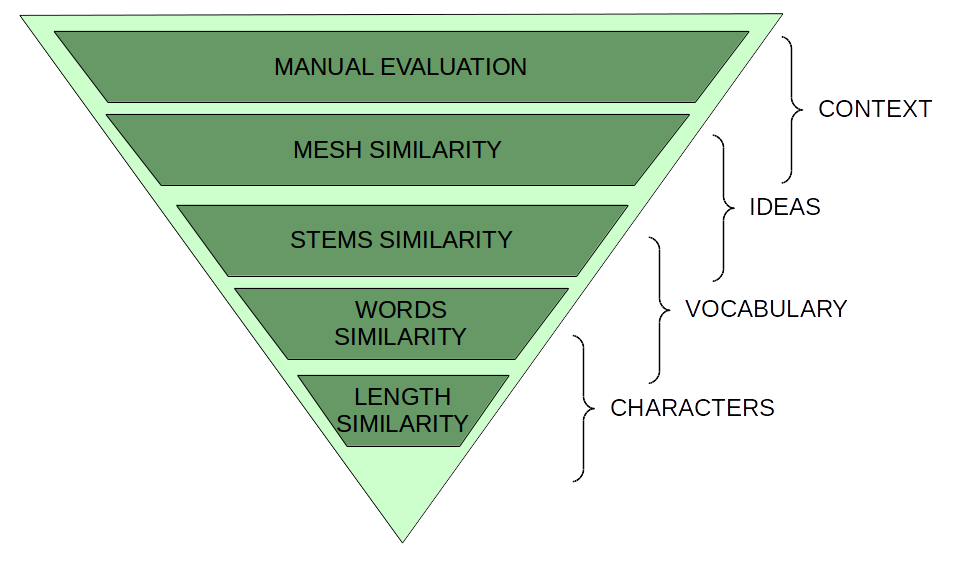}
 \caption{\textbf{Ranking of the five designed documents similarity evaluation tasks.} These tasks aim to cover every level of similarity, from the context to the characters through main ideas and vocabulary.}
  \label{FIGURE_TRIANGLE_EVALUATIONS}
 \end{figure}

The goal here being to assess if D2V could effectively replace the related-document function on PubMed, five different document similarity evaluations were designed as seen on figure \ref{FIGURE_TRIANGLE_EVALUATIONS}. These tasks were designed to cover every similarities, from the most general (the context) to the character-level similarity. 

Indeed, a reliable algorithm to find related documents should be able to bring closer texts sharing either a similar context, some important ideas (stems of words), an amount of non-stemmed vocabulary (\textit{e.g.} verbs tenses are taken in account) and should not be based on raw character-similarity (two documents sharing the same proportion of letter “A” or having a similar length should not be brought together if they do not exhibit upper levels similarity).

\subsubsection{String length}

To assess whether a similar length could lead to convergence of two documents, the size of the query document $D_{x}$ has been compared with the top-close document $C_{x}$ for 10,000 document randomly selected from the \textbf{TeS} after some pre-processing steps (stopwords and spaces were removed from both documents).

\subsubsection{Words co-occurrences}

A matrix of words co-occurrence was constructed on the total corpus from PubMed. Briefly, each document was lowered and tokenized. A matrix was filled with the number of times that two words co-occur in a single document. Then, for 5,000 documents $D_{x}$ from the \textbf{TeS}, all models were queried for the top-close document $C_{x}$. All possible combinations between all words $WD_{x} \in D_{x}$ and all words $WC_{x} \in C_{x}$ (excluding stopwords) were extracted, 500 couples were randomly selected and the number of times each of them was co-occurring was extracted from the matrix. The average value of this list was calculated, reflecting the proximity between D and C regarding their words content. This score was also calculated between each $D_{x}$ and the top-close document $C_{x}$ returned by the \textit{pmra} algorithm.

\subsubsection{Stems co-occurrences}

The evaluation task explained above was also applied on 10,000 stemmed texts (using the Gensim’s PorterStemmer to only keep word’s roots). The influence of the conjugation form or other suffixes can be assessed.

\subsubsection{MeSH similarity}

It is possible to compare the ability of both \textit{pmra} and D2V to bring closer articles which were indexed with common labels. To do so, 5,000 documents $D_{x}$ randomly selected from the \textbf{TeS} were sent to both \textit{pmra} and D2V architectures, and the top-five closer articles $C_{x}$ were extracted. The following rules were then applied to each MeSH found associated with $D_{x}$ for each document $C_{x_i}$ : add 1 to the score if this MeSH term is found in both $D_{x}$ and $C_{x_i}$, add 3 if this MeSH is defined as major topic and add 1 for each qualifier in common between $D_{x}$ and Cxi regarding this particular MeSH term. Then, the mean of these five scores was calculated for both \textit{pmra} and D2V.

\subsubsection{Manual evaluation}

Among all documents contained in the \textbf{TeS}, 10 articles $D_{x}$ have been randomly selected. All of them were sent to the \textit{pmra} and to the most accurate of the two D2V architectures, regarding the automatic evaluations explained above. Each model was then queried for the ten closest articles for each $D_{x_i} \in D_{x}$ and the relevance between $D_{x_i}$ and every of the top-ten documents was blindly assessed by a three-modality scale used in other standard Information Retrieval test sets: bad (0), partial (1) or full relevance (2) \cite{hersh1994ohsumed}. In addition, evaluators have been asked to rank publications according their relevant proximity with the query, the first being the closest from their perspective. Two medical doctors and two medical data librarians took part in this evaluation.

\section{Results}

\subsection{Optimisation}

Regarding the optimisation, 1,920 different models were trained and evaluated. First, the \textit{dm} parameter highly affects the accuracy. Indeed, the PV-DBOW architecture looks more precise with a highest accuracy of 25.78\%, while the PV-DM reached only 18.08\% of common MeSH terms in average between query and top-close documents. Then, embedding vectors having large number of dimensions ($> 256$) seem to lead to a better accuracy, for PV-DBOW at least. Finally, when set too low ($< 0.01$), the \textit{alpha} parameter leads to poor accuracy. The best combination of parameters, obtained thanks to the PV-DBOW architecture, was selected. The best parameters regarding the PV-DM, but having the same \textit{vector\_size} value, were also kept (13.30\% of accuracy). The concatenation of models is thus possible without dimensions reduction, this method being promoted by Mikolov and Lee \cite{le2014distributed}. Selected values are listed on the table \ref{table_selected_parameters}.

\begin{table}
\caption{Parameters leading to the higher accuracy during the optimisation task.}
\label{table_selected_parameters}
      \begin{tabular}{cccccc}
        \hline
        dm & vector\_size  & sample  & alpha & window & hs\\ \hline
        0 & 512 & 0.0001 & 0.01 & 9 & 1\\
        1 & 512 & 0.00001 & 0.1 & 5 & 0\\ \hline
      \end{tabular}
\end{table}

\subsection{Evaluation}

\subsubsection{String length}

\begin{figure}
 \includegraphics[scale=0.15]{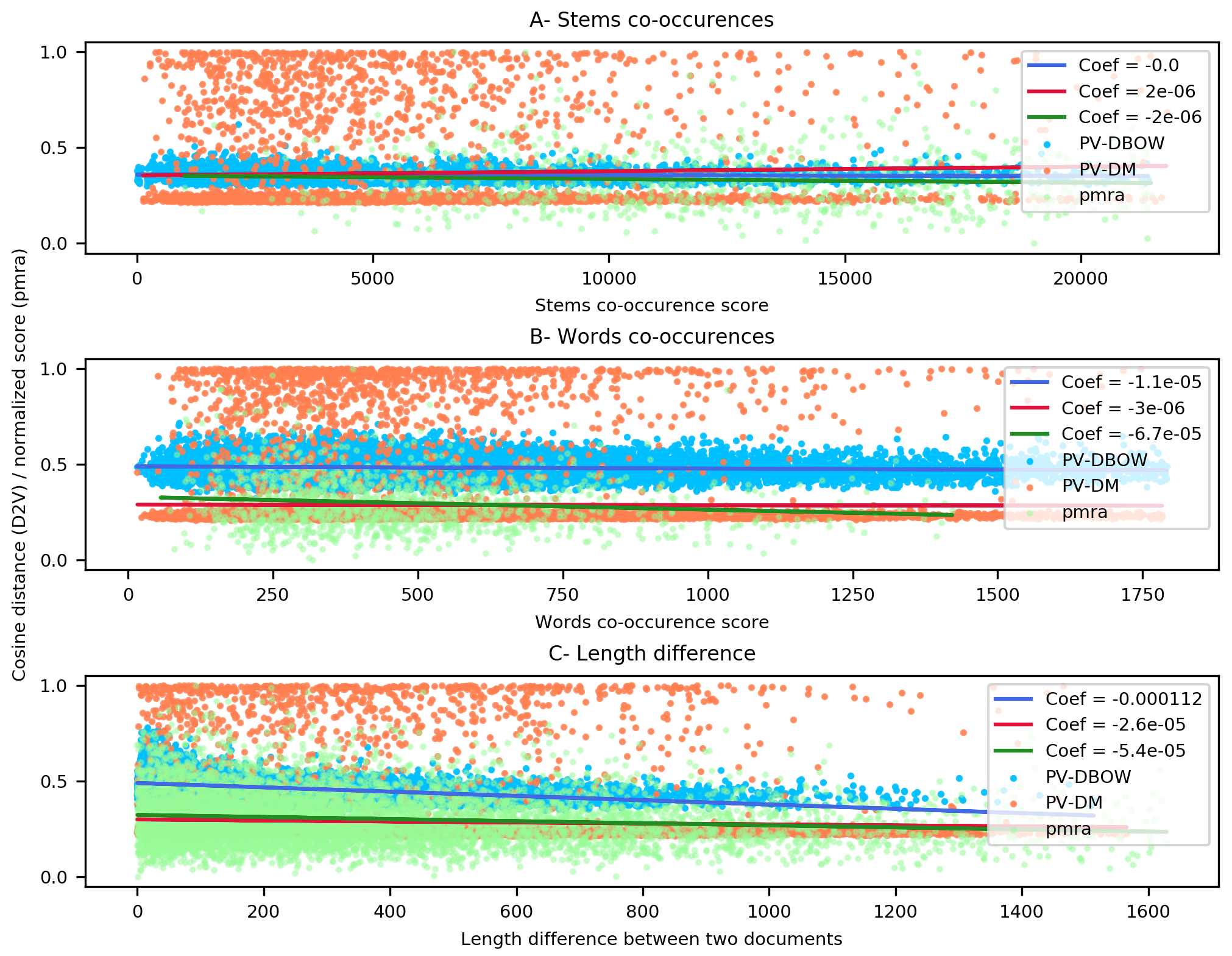}
  \caption{\textbf{Analysis of stems, words and length differences between texts broughts closer by D2V and \textit{pmra}.} Correlation plot between the stems co-occurrence score (A), words co-occurrence score (B), length difference (C) and scores returned by two D2V architectures (PV-DBOW, blue and PV-DM, orange) or the \textit{pmra} model (green, normalized values). Outliers with z-score > 3 were discarded from the plot.}
  \label{figure_stems_words_length}
 \end{figure}

By looking at the length difference in term of characters between documents brought closer by D2V, a difference is visible between the two architectures (Figure \ref{figure_stems_words_length}C). In fact, while a very low correlation is visible under the PV-DM architecture (coefficient $-2.6e10^{-5}$) and under the \textit{pmra} model ($-5.4e10^{-5}$), a stronger negative one is observed between the cosine distance computed by the PV-DBOW for two documents and their difference in terms of length (coefficient $-1.1e10^{-4}$). This correlation suggests that two documents having a similar size are more likely to be closer in the vectorial space created by the PV-DBOW (cosine distance closer to 1).

\subsubsection{Words co-occurrences}

Once scores from \textit{pmra} have been normalized, the correlation between words co-occurrences and scores returned by both D2V and \textit{pmra} were studied (Figure \ref{figure_stems_words_length}B). The very low slopes of the D2V trend lines ($-1.1e10^{-5}$ for the PV-DBOW and $-3e10^{-6}$ for PV-DM) indicate that the vocabulary content does not influence (positively or negatively) the proximity between two documents for this algorithm. By looking at the green dots or line, the \textit{pmra} seems to give less importance to the co-occurrence of terms. A low slope is observed ($-5.8e10^{-5}$), indicating a slight negative correlation between word co-occurrence and computed score.

\subsubsection{Stems co-occurrences}

This test assigns a score reflecting the proximity between two documents regarding their vocabulary content, the impact of the conjugation, plural forms, etc was lowered by a stemming step. The D2V model returns a cosine score S for a pair of documents ($0 < S < 1$, the top-close document is not likely to have a negative cosine value), while the \textit{pmra} returns a score between 18M and 75M in our case \cite{sayers2009utilities}. These scores were normalized to fit between the same limits than the cosine distance. For PV-DBOW, PV-DM and \textit{pmra}, the influence of the stems is almost insignificant with very flat slopes looking at the trend lines ($1e10^{-6}$, $-2e10^{-6}$ and $-2e10^{-6}$ respectively, see figure \ref{figure_stems_words_length}A). This indicates that the stem content of two documents will not affect (negatively or positively) their proximity for these models.

\subsubsection{MeSH similarity}

\begin{figure}
 \includegraphics[scale=0.15]{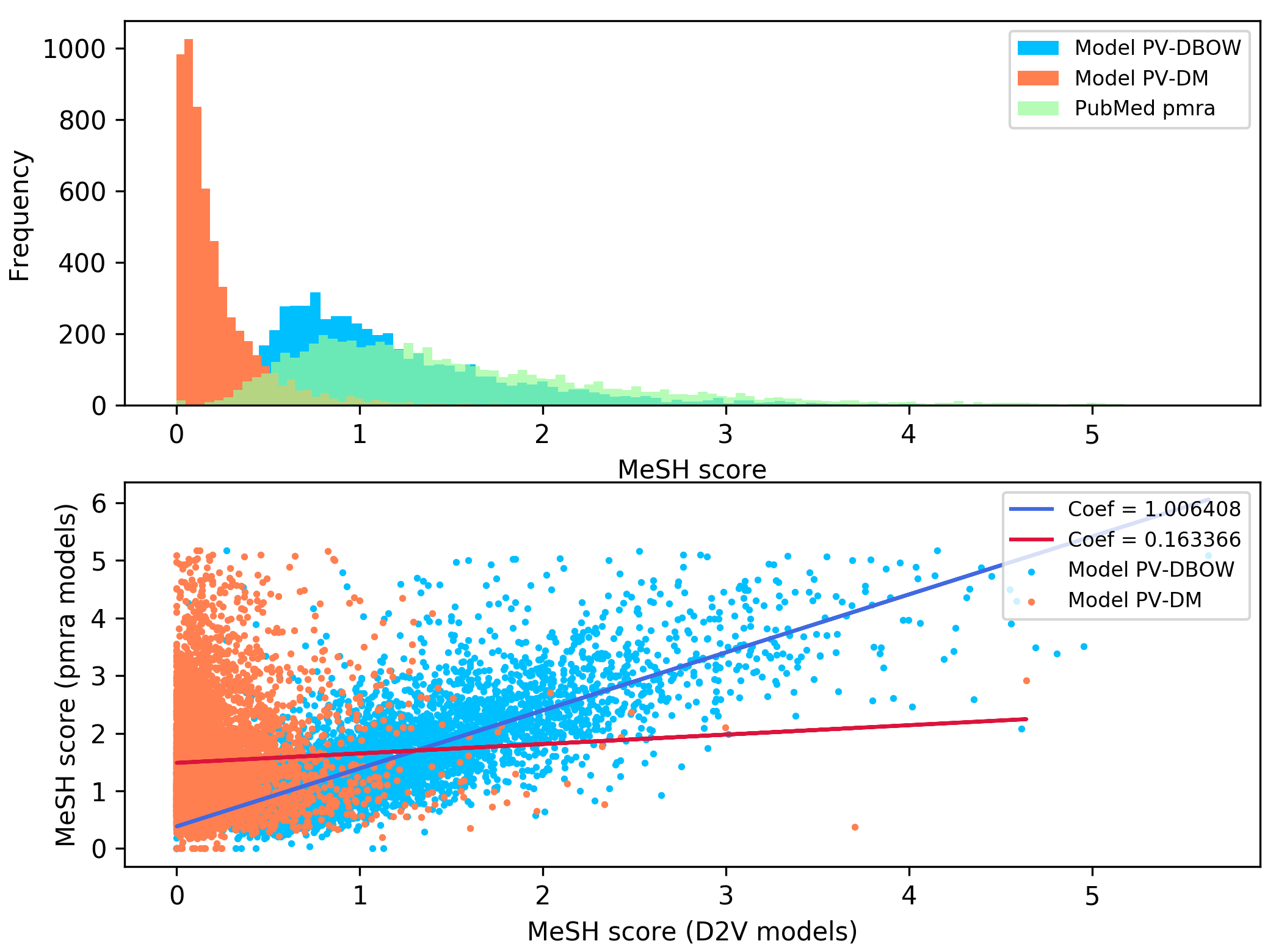}
  \caption{\textbf{Study of both \textit{pmra} and D2V models regarding their ability to bring closer documents sharing many MeSH labels.} A (upper panel): frequency of the different MeSH scores for the \textit{pmra}, PV-DM and PV-DBOW models. PV-DBOW and \textit{pmra} are centred on the same value and have a similar distribution, indicating a common ability to link documents regarding their topic. However, the PV-DM algorithm looks less efficient. B (lower panel): correlation between MeSH scores calculated from the \textit{pmra} and those from D2V. The slopes of the trend lines support the precedent result with a slope close to 1 for PV-DBOW while the PV-DM only reach 0.1, indicating a weaker correlation. Outliers with z-score > 3 were discarded from the plot.}
  \label{figure_mesh_evaluation}
  \end{figure}

By studying the common MeSH labels between two close documents, it is possible to assess whether the context influence or not this proximity. By looking at the figure \ref{figure_mesh_evaluation}A, we can see that PV-DBOW and \textit{pmra} are very close in term of MeSH score, indicating that they bring closer documents sharing a similar number of common MeSH labels in average. The \textit{pmra} model seems to be more likely to output documents sharing a higher MeSH score (the distribution tail going further 4 with a mean equal to 1.58, standard deviation: 1.06), while the PV-DM brings closer documents that are less likely to share an important number of MeSH terms, with a majority of score between 0 and 1 (mean equal to 1.16, standard deviation: 0.73). The figure \ref{figure_mesh_evaluation}B shows the correlation between the MeSH score for documents returned by the \textit{pmra} and those returned by both PV-DM and PV-DBOW models. The PV-DBOW algorithm looks way closer to the \textit{pmra} in terms of common MeSH labels between two close documents with a slope of 1.0064. The PV-DM model is much less correlated, with a slope of 0.1633, indicating less MeSH in common for close articles.

\subsubsection{Manual evaluation}

Regarding the results obtained by both PV-DBOW and PV-DM sub-architectures, the PV-DBOW model has been used versus the \textit{pmra}. Its close score in the MeSH evaluation task compared to the \textit{pmra}'s one indicates an ability to bring closer documents sharing same concepts. Thus, 10 randomly chosen documents were sent to the \textit{pmra} and to the PV-DBOW models and they were asked to output the 10 closest documents for each. Their relevance was then assessed by four evaluators.

\begin{figure}
 \includegraphics[scale=0.3]{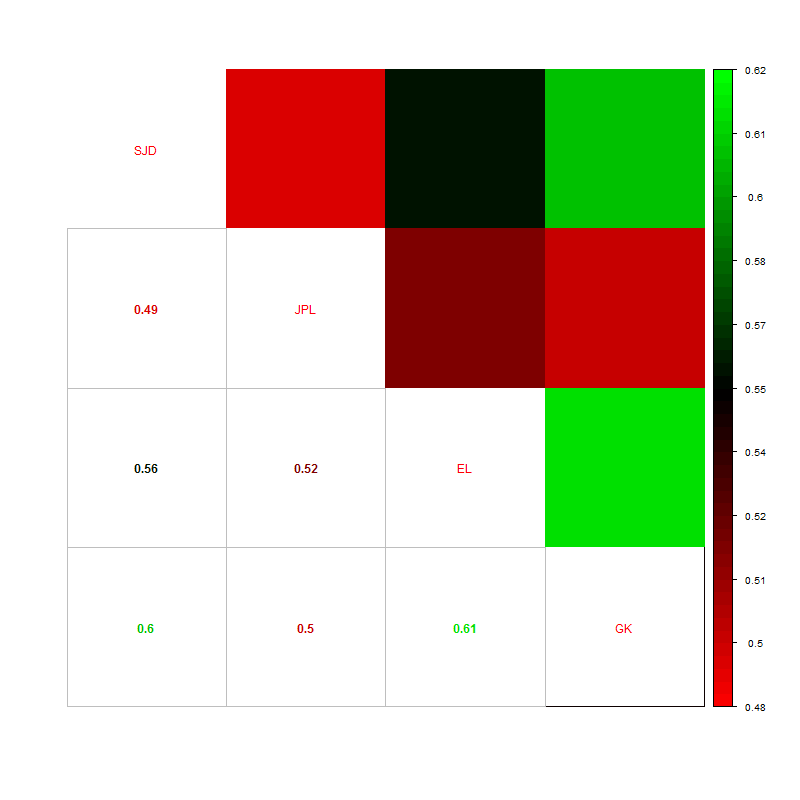}
  \caption{\textbf{Global agreement between four evaluators rating the accuracy of the D2V and \textit{pmra} models.} Colour scale indicates the strength of the agreement between two annotators. It ranges from 0.49 between the two medical doctors SJD and JPL to 0.61 between the two medical data librarian EL and GK.}
  \label{figure_manual_evaluation}
  \end{figure}

The agreement between all evaluators regarding the three-modalities scale was assessed by computing the Cohen's kappa score $K$ thanks to the SKlearn Python's library (Figure \ref{figure_manual_evaluation}) \cite{scikit-learn}. First, we can notice that the highest $K$ was obtained by the two medical data librarian (EL and GK) with $K=0.61$, indicating a substantial agreement \cite{cohen1968weighted}. In contrary, the lowest $K$ was computed using evaluations from the two Medical Doctors (SJD and JPL) with $K=0.49$, indicating barely a moderate agreement. The average agreement is represented by $K=0.55$, indicating a moderate global agreement.

Regarding the ranking of all results (the first being the most accurate compared to the query, the last the worst one), the agreement can also be seen as moderate. The concordance rate has been defined between two evaluators for a given pair of results $A/B$ as the probability for A to be better ranked than B for both judges. For each couple of evaluators the mean agreement was computed by averaging ten pairs $result/query$ randomly selected. In order to evaluate the 95\% bilateral confidence interval associated with the average concordance rate of each pair of judges the Student confidence interval estimation method has been used. Deviation from normal has been reduced by  hyperbolic arc-tangent transformation. The global mean concordance by pooling all judges together was 0.751 (\textit{sd} = 0.08). The minimal concordance was equal to 0.73 and the maximal one to 0.88.

\begin{figure}
 \includegraphics[scale=0.3]{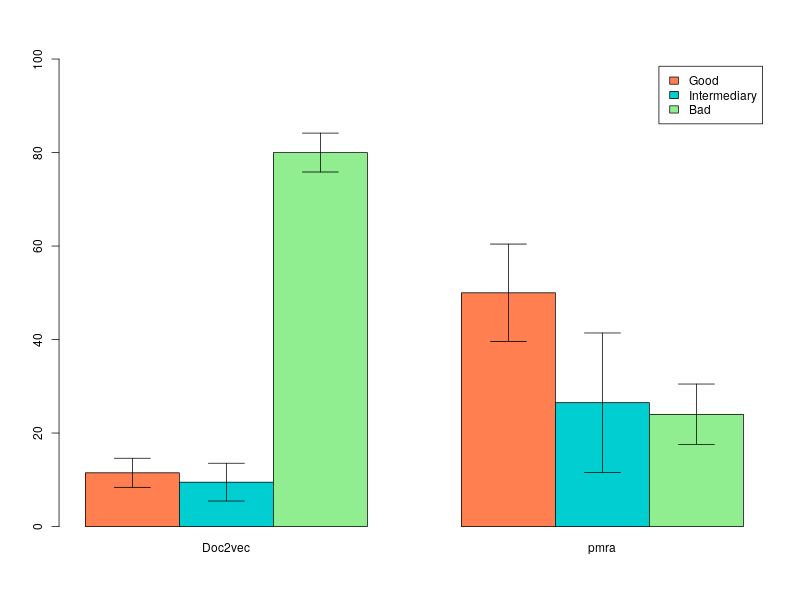}
  \caption{\textbf{Pulled rating of both models D2V and \textit{pmra}.} The height indicates the number of times each model has been rated as \textit{bad}, \textit{moderate} or \textit{strong} accuracy result by the evaluators. D2V has been mostly rated as \textit{badly} relevant (80 times) while the \textit{pmra} was mostly rated as \textit{good} relevance.}
  \label{figure_manual_evaluation_scores}
  \end{figure}

Regarding the evaluation itself, based on the three-modality scale (bad, partial or full relevance), models are clearly not equivalents (Figure \ref{figure_manual_evaluation_scores}). The D2V model has been rated 80 times as "bad relevance" while the \textit{pmra} returned only 24 times badly relevant documents. By looking at the results ranking, the mean position for D2V was 14.09 (ranging from 13.98 for JPL to 14.20 for EL). Regarding the \textit{pmra}, this average position was equal to 6.89 (ranging from 6.47 for EL to 7.23 for SJD).

\section{Discussion}

In this study, the ability of D2V to infer similarity between biomedical abstracts has been compared versus the \textit{pmra}, the algorithm actually used in Pubmed. 


Regarding the strings length task, even if trending lines slopes are very close to zero, a slight negative correlation is observed between the difference in terms of character and scores calculated by PV-DBOW and \textit{pmra}. This result can be relativized. Indeed, it was expected that two different abstracts regarding their number of characters are more likely to be different in term of context. The longest text can treat more subjects with different words (explaining D2V’s results) or to be associated with more MeSH labels (clarifying \textit{pmra} ones’).


Words or stems content analysis does not showed any particular correlation between common words/stems and scores computed by both D2V models or \textit{pmra}. Inverse results could have been expected, regarding the way \textit{pmra} is linking documents (using common terms between documents). The score brought to the \textit{pmra} model by the MeSH terms should be quite important for the final scoring formula. However, among all possible couples of words between two documents, only 500 were randomly selected, due to computational limits. Random sampling effect could have led to these results.


D2V takes in account many language features such as bi- or trigrams, synonyms, other related meanings and stopwords. No prior knowledge of analysis on the documents are needed. The \textit{pmra} is based (in addition to words) on the manual MeSH indexing of the document, even if this aspect was not discussed in the Lin and Wilbur’s publication. This indexing step is highly time-consuming and employs more than 50 people to assign labels on documents from PubMed. The result displayed on the figure \ref{figure_mesh_evaluation} could have been expected for the \textit{pmra} algorithm, this model using the MeSH terms on the statistical formula used to link documents as well as \textit{elite} or \textit{elitness} terms. It was thus expected that two documents sharing a lot of indexing labels would have been seen close by the \textit{pmra}. However, these MeSH descriptors were only used to select the appropriate parameters used to train the D2V models. The fact that D2V still manages, with the PV-DBOW architecture, to find documents that are close to each other regarding the MeSH indexing demonstrates its ability to capture an article’s subject solely with its abstract and title.


Regarding the manual evaluation, D2V PV-DBOW model has been very largely underrated compared to the \textit{pmra} model. Its results have been seen as not accurate more than three times compared to the Pubmed's model. Regarding the ranking of the results, the average position of the \textit{pmra} is centred around 7, while D2V's one is around 14. However, the real signification of these results can be relativised. Indeed, the agreement between the four annotators is only moderate and no general consensus can be extracted. 


This study also has some limitations. First, the MeSH indexing of documents on PubMed can occur on full-text data, while both optimisation of the hyper-parameters and an evaluation task are based on abstracts' indexing. However, this bias should have a limited impact on the results. The indexing being based on the main topics from the documents, these subjects should also be cited in the abstract. About this manual indexing, a bias is brought by the indexers. It is well-known in the information retrieval community that intra- and inter-indexers bias exist.

As the parameters optimisation step relied only on MeSH terms, it assumed that a model trained on articles’ abstracts can be optimised with MeSH terms which are selected according to the full text of the articles. In other words, this optimisation assumed an abstract is enough to semantically represent the whole text. But this is not completely true. If it was, MeSH terms would have not be selected on full texts in the first place. Also, the principle that a PubMed related article feature has to give articles which have a lot of MeSH terms in common has been followed throughout this work. 


To go further, as mentioned in the paper presenting D2V, the concatenation of vectors from both PV-DM and PV-DBOW for a single document could lead to a better accuracy. A third model could be designed by the merge of the two presented here. Another moot point on the text embedding community is about the part-of-speech tagging of the text before sending it to the model (during both training and utilisation). This supplementary information could lead to a better understanding of the text, particularly due to the disambiguation of homonyms.

\section{Conclusion}

This study showed that Doc2Vec PV-DBOW, an unsupervised text embedding technique, can infer similarity between biomedical articles' abstract. It requires no prior knowledge on the documents such as text indexing and is not impacted by raw words content or document structure. This algorithm was able to link documents sharing MeSH labels in a similar way the \textit{pmra} did. A manual evaluation returned very low scores for the D2V PV-DBOW model, but with a highly moderate agreement between evaluators. More investigation should be carried out to understand this difference between the evaluation based on the MeSH indexing (performed by humans) and the manual evaluation.

\bibliographystyle{unsrt}  
\bibliography{main}
\end{document}